\documentclass[a4paper]{article}

\usepackage{INTERSPEECH_v2}
\usepackage{amsmath}

\title{Combining Residual Networks with LSTMs for Lipreading}
\name{Themos Stafylakis, Georgios Tzimiropoulos}
\address{Computer Vision Laboratory\\
University of Nottingham, UK}
\email{\{themos.stafylakis, yorgos.tzimiropoulos\}@nottingham.ac.uk}

\begin{document}

\maketitle
\begin{abstract}
We propose an end-to-end deep learning architecture for word-level visual speech recognition. The system is a combination of spatiotemporal convolutional, residual and bidirectional Long Short-Term Memory networks. We train and evaluate it on the Lipreading In-The-Wild benchmark, a challenging database of 500-size target-words consisting of 1.28sec video excerpts from BBC TV broadcasts. The proposed network attains word accuracy equal to 83.0\%, yielding 6.8\% absolute improvement over the current state-of-the-art, without using information about word boundaries during training or testing.  
\end{abstract}

\section{Introduction}

Visual speech recognition (also known as lipreading) is a field of growing attention. It is a natural complement to audio-based speech recognition that can facilitate dictation in noisy environments and enable silent dictation in offices and public spaces. It is also useful in applications related to improved hearing aids and biometric authentication, \cite{assael2016lipnet}. Lipreading is the field where the speech recognition and computer vision communities meet each other and combine the advances of each field. The tremendous success of deep learning in both fields has already affected visual speech recognition, by shifting the research direction from handcrafted features and HMM-based models to deep feature extractors and end-to-end deep architectures. Recently introduced deep learning systems beat human lipreading experts by a large margin, at least for the constrained vocabulary defined by each database,  \cite{assael2016lipnet} \cite{chung2016lipsent}.    

One way to categorize visual and audio-visual speech recognition approaches is (i) to those that model words (e.g. \cite{wand2016lipreading} \cite{chung2016lip}) and (ii) to those that model visemes (e.g. \cite{assael2016lipnet} \cite{chung2016lipsent}), i.e. visual units that correspond to sets of visually indistinguishable phonemes, \cite{fisher1968confusions} \cite{bear2016decoding}. The former approach is considered more pertinent to tasks like isolated word recognition, classification and detection, while the latter to sentence-level classification and large vocabulary continuous speech recognition (LVCSR). Nevertheless, recent advances in speech recognition and natural language processing show that direct modeling of words is feasible even for LVCSR, \cite{bengio2014word} \cite{soltau2016neural} \cite{audhkhasi2017direct}.      

The proposed system belongs to the former category, although it can support viseme-level recognition by using viseme instead of word labels at the SoftMax layer. It combined three sub-networks: (i) The front-end, which applies spatiotemporal convolution to the frame sequence, (ii) a Residual Network (ResNet) that is applied to each time step, and (iii) the back-end, which is a two-layer Bidirectional Long Short-Term Memory (Bi-LSTM) network. The SoftMax layer is applied to all time steps and the overall loss is the aggregation of the per time step losses, and the system is trained in an end-to-end fashion. Finally, the system performs not merely word recognition but also implicit key-word spotting, since the target words are not isolated, but they are part of whole utterances of fixed duration (1.28sec). Information regarding word boundaries is not utilized neither during training nor during evaluation.\footnote{Code and pre-trained models in Torch7 are available at https://github.com/tstafylakis/Lipreading-ResNet}

The rest of the paper is organized as follows. In Section \ref{Related_Work}, we refer to recent works on visual speech recognition, with emphasis on those that apply deep learning methods. The Lipreading In-The-Wild (LRW) database is discussed in Section \ref{Database}, while in Section \ref{Our_Model} we present analytically the proposed model, together with some useful detail about preprocessing and implementation. Finally, in Section \ref{Exper} we present our experimental results, together with baseline and state-of-the-art results.
\section{Related work}
\label{Related_Work}
Prior to the advent of deep learning (\cite{hinton2012deep}) most of the work in lipreading was based on hand-engineered features, that were usually modeled by HMM-based pipeline, \cite{goldschen1997continuous} \cite{chiou1997lipreading} \cite{potamianos2003recent} \cite{chandrasekaran2009natural} \cite{papandreou2009adaptive}. Spatiotemporal descriptors such as active appearance models and optical flow, and SVM classifiers have also been proposed, \cite{shaikh2010lip}. For an analytic review on traditional lipreading methods we refer to \cite{zhou2014review} and \cite{potamianos2004audio}. More recent works deploy deep learning methods either for extracting "deep" features (\cite{noda2015audio} \cite{thangthai2015improving} \cite{almajai2016improved}) or for building end-to-end architectures. In \cite{huang2013audio}, Deep Belief Networks were deployed for audio-visual recognition and 21\% relative improvement was reported over a baseline multi-stream audio-visual GMM/HMM system. In \cite{petridis2016deep}, bottleneck features are extracted using Deep Autoencoder. The bottleneck features are concatenated with DCT features and the overall system is trained jointly using an LSTM back-end. In \cite{wand2016lipreading}, a fully LSTM architecture is proposed, which attains superior results compared to traditional methods on the GRID audiovisual corpus, \cite{GRID}. In \cite{assael2016lipnet}, an end-to-end sentence-level lipreading network (LipNet) is introduced, that combines spatiotemporal convolutional layers, LSTMs and Connectionist Temporal Classification (CTC, \cite{graves2006connectionist}). It attains 95.2\% sentence-level accuracy on a subset of speakers from GRID database, while trained on the remaining GRID speakers. Finally, in \cite{chung2016lipsent}, the encoder-decoder with attention mechanism is explored, in both audio-visual and visual settings. Using solely visual information, 97.0\% word accuracy is reported on GRID and 76.2\% word accuracy on LRW. To the best of our knowledge, the latter results define the current state-of-the-art for both databases, insofar as additional training resources may be leveraged, \cite{chung2016lipsent}.   
\section{Database}
\label{Database}

We train and evaluate the algorithm on the challenging LRW database, \cite{chung2016lip}. The database consists of audiovisual speech segments extracted from BBC TV broadcasts (News, Talk Shows, a.o.) and it is characterized by its high variability with respect to speakers and pose. Moreover, the number of target words is 500, which is an order of magnitude higher than other publicly available databases (GRID \cite{GRID}, CUAVE \cite{CUAVE}, a.o.). Another feature that renders the database so challenging is the existence of pairs of words that share most of their visemes. Such examples are nouns in both singular and plural forms (e.g. benefit-benefits, 23 pairs) as well as verbs in both present and past tenses (e.g. allow-allowed, 4 pairs). 

However, perhaps the most difficult aspect of the database -and of the setting we chose to proceed with- is the fact that the target-words appear within utterances rather than being isolated. Hence, the network should learn not merely how to discriminate between 500 target-words, but also how to ignore the irrelevant parts of the utterance and spot one of the target-words. And it should learn how to do so without knowing the word boundaries. Some random examples of utterances are ``...the {\em election} victory...", ``...the day's {\em other} news...", ``...and so {\em senior} labour..."  and ``...point, I {\em think} the...", where italics denote the target-word of each utterance.

\begin{figure}
\centering
\includegraphics[width=3.00in]{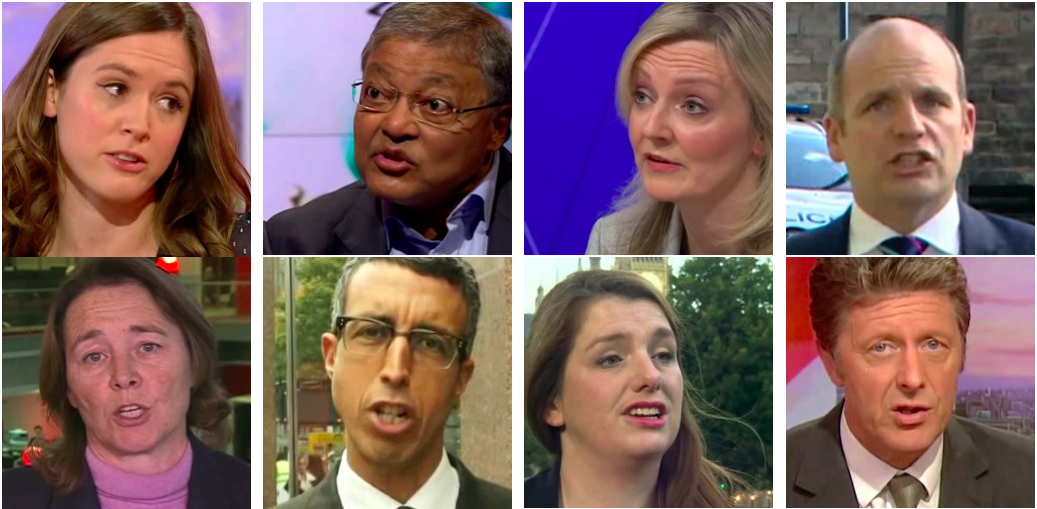}
\vspace{-1mm}\caption{Random frames from the LRW database}
\label{LRW}
\end{figure}

The collection of the database was fully automatic, involving OCR on the subtitles, synchronization with the audio (forced alignment), as well as verification that the speaker is visible (see \cite{chung2016lip} for a detailed description). The training set consists of up to 1000 occurrences per target word, while the validation and evaluation sets both consist of 50 occurrences per word. Each clip is of fixed duration (1.28sec, 31 frames with 25fps frame rate). Random frames from the database are depicted in Fig. \ref{LRW}. 

\section{Deep Learning modeling and preprocessing}
\label{Our_Model}
\subsection{Facial landmarks and data augmentation}
In the first preprocessing step, we discard redundant information in order to focus on the mouth region. To do so, we use the 2D version of the algorithm proposed in \cite{bulat2016convolutional} and \cite{bulat2016two}. The algorithm tackles regression in two steps. It first applies detection to extract a set of heatmaps (one per landmark) which are used as side information for the subsequent regression network. Based on the 66 facial landmarks, we crop the images and resize them to a fixed 112$\times$112 size. A common cropping is applied to all frames of a given clip, using the median coordinates of each landmark. 
The frames are transformed to grayscale and are normalized with respect to the overall mean and variance. Finally, data augmentation is performed during training, by applying random cropping ($\pm$5 pixels) and horizontal flips, that are common across all frames of a given clip. 
\subsection{Spatiotemporal front-end}
The first set of layers applies spatiotemporal convolution to the preprocessed frame stream. Spatiotemporal convolutional layers are capable of capturing the short-term dynamics of the mouth region and are proven to be advantageous, even when recurrent networks are deployed for back-end, \cite{assael2016lipnet}. They consist of a convolutional layer with 64 3-dimensional (3D) kernels of 5$\times$7$\times$7 size (time/width/height), followed by Batch Normalization (BN, \cite{ioffe2015batch}) and Rectified Linear Units (ReLU). The extracted feature maps are passed through a spatiotemporal max-pooling layer, which drops the spatial size of the 3D feature maps. The number of parameters of the spatiotemporal front-end is $\sim$16K.  

\subsection{Residual Network}
The 3D features maps are passed through a residual network (ResNet, \cite{he2016deep}), one per time-step. We use the 34-layer identity-mapping version, which was proposed for ImageNet, \cite{he2016identity}. Its building blocks are composed of two convolutional layers, and with BN and ReLU, while the skip connections facilitate information propagation, \cite{he2016identity}. The ResNet drops progressively the spatial dimensionality with max pooling layers, until its output becomes a single dimensional tensor per time step. We should emphasize that we did not make use of pretrained models, as they are optimized for completely different tasks (e.g. static colored images from ImageNet or CIFAR). The number of parameters of the ResNet is $\sim$21M. 
\subsection{Bidirectional LSTM back-end and optimization criterion}
The back-end of the model is a Bidirectional LSTM network. For each of the two directions, we stack two LSTMs, and the outputs of the final LSTMs are concatenated. The number of parameters of the LSTM back-end is $\sim$2.4M. 

When using word-level recognition without explicit modelling of visemes, several approaches exist in terms of the optimization criterion. One approach is to place the SoftMax layer at the last time step of the LSTM output, i.e. when the overall sequence is encoded by the LSTM. Backpropagation through time is capable of propagating the errors all the way back to the first time step of the sequence, given the resilience of LSTM to the problem of vanishing gradients, \cite{wand2016lipreading}. A second approach is to apply the criterion for each time step. This approach is closer to the typical use of LSTMs in speech recognition, where instead of phoneme/viseme labels, the word label is repeated at every time step. This approach fits well to bidirectional LSTMs, since hidden states have in all time steps access to the overall video, \cite{graves2005bidirectional}. After experimentation with both approaches, we concluded that the latter leads to much higher word accuracy (about 3\% absolute improvement). Hence, the overall loss is defined as the aggregated loss over all time steps, which coincides to the summation of negative logarithm of word posteriors. Notice again that the word label is applied to all time steps of the clip, since word boundaries are unknown. 
\begin{figure}[!htbp]
\centering
\includegraphics[width=2.80in]{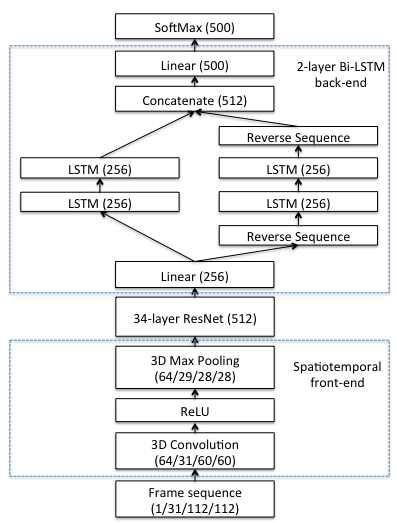}
\vspace{-2mm}\caption{The block-diagram of the proposed network.}
\label{diagram}
\end{figure}
\subsection{Implementation details}
Our implementation is based on Torch7 (\cite{torch7}) and the networks are trained on a NVIDIA Titan X (Pascal Architecture) GPU with 12GB memory. We use the standard SGD training algorithm, with momentum 0.9. BN follows all convolutional and linear layers, apart from the one preceding the SoftMax layer. We do not apply dropout, since it is not part of the ResNet training recipe (BN seems to suffice, \cite{ioffe2015batch}). The initial learning rate is $5\times10^{-3}$ for the experiments with the convolutional backend and $5\times10^{-4}$ for those with Bi-LSTM, while the final is $5\times10^{-5}$, decreasing on log scale. Training is considered complete when the results on the validation set do no longer improve, with a delay of 3 epochs. All our models converge after 15 to 20 epoches. 

A block-diagram of the network is depicted in Fig. \ref{diagram}. BN layers have been omitted for clarity. The size of the tensors each layer outputs is also presented. For the 3D-convolutional front-end, tensor dimensions denote channels, time, width and height. 

We should emphasize that although the overall system can be directly trained end-to-end, we use the following three steps approach. Initially, a temporal convolutional back-end is used instead of the Bi-LSTM. After convergence, the temporal convolutional back-end is removed and the Bi-LSTM back-end is attached. The Bi-LSTM is trained for 5 epochs, keeping the weights of the 3D convolution front-end and the ResNet fixed. Finally, the overall system is trained end-to-end. A comparison between the two back-ends is presented in Section \ref{Exper}.      

\section{Experiments}
\label{Exper}
\subsection{Baseline results}
The best baseline result published in \cite{chung2016lip} is the multi-tower VGG-M. It consists of a set of parallel VGG models (towers) with shared weights, which are concatenated channel-wise using pooling, while the rest of the network is the same as the regular VGG-M. The results are presented in Table \ref{baseline} in terms of word accuracy. Top-1 corresponds to the percentage of times the word was correctly identified, while more generally Top-$N$ corresponds to the percentage of times the correct word was among the $N$ best scores.

\begin{table}[!htbp]
\centering
\begin{tabular}{| c || c | c | c |} 
\hline
Network & Top-1 & Top-5 & Top-10 \\ [0.5ex] 
\hline
Baseline & 61.1\% & - & 90.4\% \\ 
\hline
\end{tabular}
\vspace{1mm}\caption{Word accuracies for the baseline network (VGG-M).}\label{baseline}
\end{table}

In \cite{chung2016lipsent}, an attentional encoder-decoder architecture is proposed, \cite{xu2015show}. It is trained on a different set of BBC TV Broadcasts, which contains whole sentences rather than words. A visual-only version of the system (termed "Watch, Attend and Spell", WAS) is evaluated on GRID and on LRW. The network is pretrained on the BBC TV Broadcasts, while the training sets of GRID and LRW are used for fine-tuning. Word accuracies (Top-1) equal to 97.0\% and 76.2\% are reported on GRID and LRW respectively, which according to our knowledge represent the current state-of-the-art on both databases.

\subsection{Results using our network}
We begin by using a simpler model than the proposed one, in order to examine the contribution of each individual component of the network. The first network applied 2D convolution instead of 3D. The 2D convolution is followed by the ResNet, while the back-end is based on temporal convolution rather than LSTMs. More specifically, we use two temporal convolutional layers, each of which is followed by BN, ReLU and Max Pooling which reduce the temporal dimensions by a factor of 2. Finally, a Mean Pooling layer is added, followed by a linear and a SoftMax layer. The results are presented in Table \ref{Temporal} (denoted by N1). The results of the same model, but with 3D convolution are also presented in Table \ref{Temporal} (denoted by N2). 

In order to verify the effectiveness of the ResNet we replace it with a Deep Neural Network (DNN) of approximately the same number of parameters ($\sim$20M). The DNN is composed of 3 fully connected hidden layers, with BN and ReLU. Its inputs are 3D convolutional maps, treated as vectors (one per time step). The DNN progressively reduces the size of the vectors as 50176 $\rightarrow$ 384 $\rightarrow$ 384 $\rightarrow$ 256. The results are presented in Table \ref{Temporal} (denoted by N3).  

\begin{table}[!htbp]
\centering
\begin{tabular}{| c || c | c | c|} 
\hline
Network & Top-1 & Top-5 & Top-10 \\ [0.5ex] 
\hline
N1 & 69.6\% & 90.4\% & 94.8\% \\ 
\hline
N2 & 74.6\% & 93.4\% & 96.5\% \\ 
\hline
N3 & 69.7\% & 90.5\% & 94.6\% \\ 
\hline
\end{tabular}
\vspace{1mm}\caption{Word accuracies using temporal convolution back-end.}\label{Temporal}
\end{table}

We now focus on the back-end of the network and use LSTMs instead of temporal convolutions. The first network in Table \ref{LSTM} (denoted by N4) uses a single-layer Bi-LSTM, while the second one (denoted by N5) uses a double-layer Bi-LSTM. These two networks are not trained end-to-end. While training the back-end, the 3D convolutional layer and the ResNet (that are copied from N2) remain fixed. Moreover, the outputs of the two directional LSTMs are added together instead of concatenated together.  

\begin{table}[!htbp]
\centering
\begin{tabular}{| c || c | c | c|} 
\hline
Network & Top-1 & Top-5 & Top-10 \\ [0.5ex] 
\hline
N4 & 78.4\% & 94.9\% & 97.4\% \\ 
\hline
N5 & 79.6\% & 95.3\% & 96.3\% \\ 
\hline
\end{tabular}
\caption{Word accuracies using different LSTM in the back-end.}\label{LSTM}
\end{table}

For the final set of results we use end-to-end training of the overall network. The first network in Table \ref{E2E} (denoted by N6) is the same as N5, but trained end-to-end, using the weights of N5 as starting point. Finally, N7 is also trained end-to-end and the sole difference with N6 is that the outputs of the two directional LSTMs are concatenated together instead of added together (as depicted in Fig. \ref{diagram}). 

\begin{table}[!htbp]
\centering
\begin{tabular}{| c || c | c | c |} 
\hline
Network & Top-1 & Top-5 & Top-10 \\ [0.5ex] 
\hline
N6 & 81.5\% & 96.0\% & 98.0\% \\ 
\hline
N7 & 83.0\% & 96.3\% & 98.3\% \\ 
\hline
\end{tabular}
\vspace{1mm}\caption{Word accuracies using LSTM back-end and end-to-end training.}\label{E2E}
\end{table}

\begin{figure}
\centering
\includegraphics[width=3.20in]{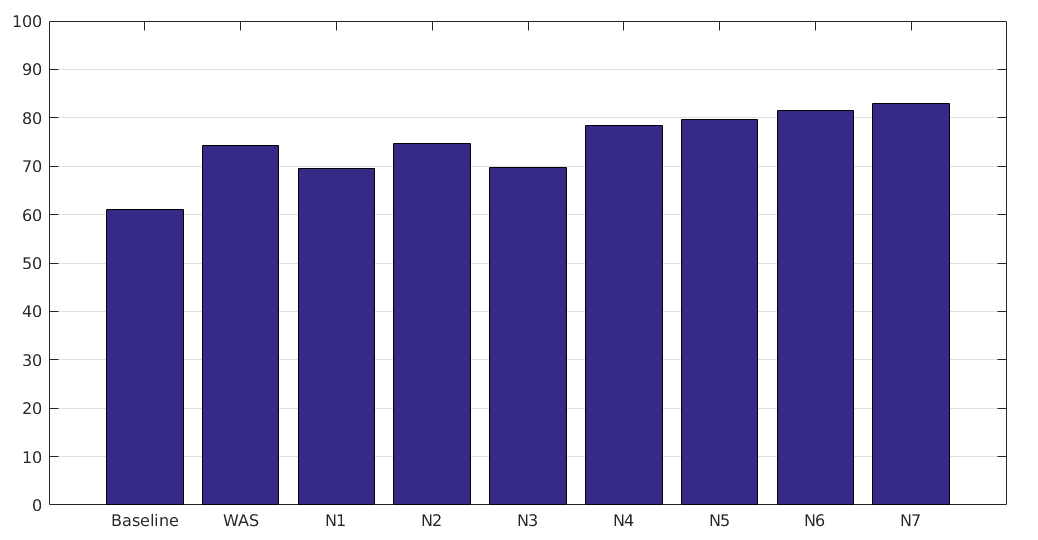}
\vspace{-5mm}\caption{Word accuracy of the networks examined.}
\label{chart}
\end{figure}

\subsection{Discussion and error analysis}
Several conclusions can be drawn from the results presented above (see also Fig. \ref{chart} for clarity). First of all, by comparing the baseline to N1, we observe our simplest system yielding 8.5\% absolute improvement over the VGG-M baseline. Moreover, the use of 3D (N2) instead of 2D (N1) leads to a further 5.0\% absolute improvement, emphasizing the need of modeling the short-term dynamics of the mouth region in the front-end. By comparing N2 and N3 we notice that the ResNet yields 4.9\% better work accuracy compared to a 3-layer DNN with the same number of parameters. In addition, by using a single-layer Bi-LSTM (N4) instead of a temporal convolutional back-end, a further 3.8\% absolute improvement is attained, highlighting the expressive power of LSTMs in modeling temporal sequences. Furthermore, the use of a two-layer Bi-LSTMs (N5) offers a further 1.2\% absolute improvement. The final set of results demonstrates the importance of end-to-end training towards achieving higher word accuracy. By training N5 in an end-to-end fashion (N6) we obtain a 1.9\% absolute improvement, while by concatenating (N7) rather than adding together (N6) the Bi-LSTM outputs we obtain our best result, a 83.0\% work accuracy. 

\begin{table}[!htbp]
\centering
\begin{tabular}{| c | c || c |}
\hline
Target Word & Decision & Error Rate (\%) \\ 
\hline\hline
SPEND & SPENT & 26 \\	
\hline
WANTS & WANTED & 18 \\	
\hline
RUSSIAN & RUSSIA & 18 \\
\hline
BENEFIT & BENEFITS & 18 \\	
\hline
BENEFITS & BENEFIT & 16 \\	
\hline
RUSSIA & RUSSIAN & 16 \\	
\hline
CANCER & AGAINST & 16 \\	
\hline
GIVING & LIVING & 16 \\	
\hline
DIFFERENCE & DIFFERENT & 14	\\
\hline
MAKES & MEANS &14	\\
\hline
\end{tabular}
\vspace{1mm}\caption{Most frequent errors made by the proposed system.} \label{EA1}
\end{table}

Table \ref{EA1} contains the most frequent errors made by our best system (N7). We observe that most of the word pairs are mutually close with respect to their phonetic and ``visemic" content. We should emphasize again that the clips contain coarticulation with preceding and succeeding words, as they are excerpted from continuous speech. Hence, correct identification of the first and last visemes of a word is occasionally hard.

The list of words for which the system yields the best and worst performance is presented in Table \ref{EA2}. As expected, the system does very well on words with rich phonetic/visemic content and vice versa. There are 8 words for which the system made no errors, and only 3 words for which the word accuracy dropped below 50\%. Recall that the number of evaluation clips is 50 per target word (i.e. 25000 clips overall). 

\begin{table}[!htbp]
\centering
\begin{tabular}{| c | c || c | c |}
\hline
Target Word & Acc (\%) & Target Word & Acc (\%) \\ 
\hline\hline
SUNSHINE & 100 & SPEND & 58 \\	
\hline
ECONOMIC & 100 & AROUND & 58 \\	
\hline
TEMPERATURES & 100 & THING & 56 \\
\hline
WESTMINSTER & 100 & THEIR & 56 \\	
\hline
POLITICIANS & 100 & UNTIL & 54 \\	
\hline
SITUATION & 100 & GETTING & 52 \\	
\hline
OBAMA & 100 & SAYING & 50 \\	
\hline
INQUIRY & 100 & THERE & 48 \\	
\hline
MINISTER & 98 & GOING & 48	\\
\hline
FAMILIES & 98 & UNDER & 42	\\
\hline
\end{tabular}
\vspace{1mm}\caption{Words with the highest accuracy (left) vs. words with the lowest accuracy (right).}\label{EA2}
\end{table}
\vspace{-2mm}

\section{Conclusions}
We proposed a spatiotemporal deep learning network for word-level visual speech recognition. The network is a stack of a 3D convolutional front-end, a ResNet and an LSTM-based back-end, and trained using an aggregated per time step loss. We chose to experiment with the LRW database, since it combines many attractive characteristics, such as large size ($\sim$500K clips), high variability in speakers, pose and illumination, non-laboratory in-the-wild conditions, and target-words as part of whole utterances rather than isolated. We explored several network configurations, and we demonstrated the importance of each building block of the network as well as the gain in performance attained by training the network end-to-end. The proposed network yielded 83.0\% work accuracy, which corresponds to less that half the error rate of the baseline VGG-M network and 6.8\% absolute improvement over the state-of-the-art 76.2\% accuracy, attained by an attentional encoder-decoder network, \cite{chung2016lipsent} \cite{chung2016lip}.        
\section{Acknowledgements}
This work has been funded by the European Commission program Horizon 2020, under grant agreement no. 706668 (Talking Heads). The views expressed in this paper are those of the authors and do not engage any official position of the funding agencies.

\bibliographystyle{IEEEtran}

\bibliography{AV}


\end{document}